\documentclass[journal]{IEEEtran}
\usepackage{xcolor,soul,framed} 
\usepackage[colorlinks=true,linkcolor=blue,citecolor=blue,urlcolor=blue]{hyperref} 
\colorlet{shadecolor}{yellow}
\usepackage[pdftex]{graphicx}
\graphicspath{{../pdf/}{../jpeg/}}
\DeclareGraphicsExtensions{.pdf,.jpeg,.png}

\usepackage[cmex10]{amsmath}
\usepackage{array}
\usepackage{mdwmath}
\usepackage{amsmath,amsfonts}
\usepackage{mdwtab}
\usepackage{amsbsy}
\usepackage{csquotes}
\usepackage{amssymb}
\usepackage{caption}

\usepackage{amsthm}
\usepackage{eqparbox}
\usepackage{url}
\usepackage{hyperref}
\begin{document}
\bstctlcite{IEEEexample:BSTcontrol}
    \title{A Causal Analysis of $\text{CO}_2$ Reduction Strategies in Electricity Markets Through Machine Learning-Driven Metalearners}
  \author{Iman Emtiazi~Naeini$^\text{a}$, Zahra Saberi$^\text{b*}$, Khadijeh Hassanzadeh$^\text{c}$

  \thanks\hrulefill
  \thanks{
  The authors declare no conflict of interest.
  
  I. Emtiazi Naeini, Z. Saberi and K.Hassanzadeh are with the Department of Computer and Controlling Systems, Isfahan Higher Education and Research Institute, the Department of Mathematical Sciences, Isfahan University of Technology, the Department of Economics, Urmia University, respectively (Email: \texttt{iman.emtiazi@gmail.com, z\_saberi@iut.ac.ir, kh.hasanzadeh@urmia.ac.ir}).
  }
  }

\maketitle

\begin{abstract}
This study employs the Causal Machine Learning (CausalML) statistical method to analyze the influence of electricity pricing policies on carbon dioxide ($\text{CO}_2$) levels in the household sector. Investigating the causality between potential outcomes and treatment effects, where changes in pricing policies are the treatment, our analysis challenges the conventional wisdom surrounding incentive-based electricity pricing. The study’s findings suggest that adopting such policies may inadvertently increase $\text{CO}_2$ intensity. Additionally, we integrate a machine learning-based meta-algorithm, reflecting a contemporary statistical approach, to enhance the depth of our causal analysis. The study conducts a comparative analysis of learners X, T, S, and R to ascertain the optimal methods based on the defined question's specified goals and contextual nuances. This research contributes valuable insights to the ongoing dialogue on sustainable development practices, emphasizing the importance of considering unintended consequences in policy formulation.
\end{abstract}


\IEEEpeerreviewmaketitle

\section{\textbf{Introduction}}

\IEEEPARstart{T}{he} dynamic fluctuations within power systems profoundly influence the quality of power supply, posing potential threats to the safety and stability of the entire power grid. As a result, thorough investigations into electrical power systems are essential both theoretically and practically. Given their substantial scale and escalating impacts, further research is warranted to deepen our understanding and address the evolving challenges inherent to these intricate systems~\cite{yang2022novel}.

The volatility in electricity prices results from combined fluctuations originating from both the supply and demand sides, significantly influenced by diverse external factors. Key contributors include variations in fuel prices, such as coal, and fluctuations in hydroelectricity production. On the supply side, critical determinants encompass fuel prices, $\text{CO}_2$  allowances, and the capacity factor of various power sources, exerting substantial influence on the dynamics of electricity prices. 

As for the demand side of the market equilibrium, climate and temperature information constitute an essential variable, given its direct influence on customer behaviour. Furthermore, Consumption trends display a temporal reliance, impacted by variables like the time of day, day of the week, season, holidays, and events. Moreover, the system's load and generation are influenced by external weather conditions, adding a dynamic dimension to their interconnection \cite{weron2014electricity} and the impact of neighboring markets \cite{lago2018forecasting}. Because of these inherent characteristics, the dynamics of electricity prices have developed into a significantly intricate terrain characterized by remarkable complexity. This complexity manifests in exceptionally volatile pricing, characterized by sudden and unpredictable peaks that can significantly impact market dynamics \cite{weron2014electricity}.

The main roles of forecasting theory and methods are underscored in advancing economic and social development. These components contribute substantively to realizing notable economic and social benefits, reducing associated costs. The strategic application of forecasting principles becomes instrumental in fostering sustainable economic growth and societal progress \cite{xiao2019novel}.

In recent decades, there has been a significant surge in research and applications dedicated to forecasting models across various research domains. The electrical power system, a crucial element within economic structures, involves various aspects, including energy production, distribution, scheduling, and usage. Given its pivotal role in societal productivity, there's a pressing need to improve forecasting techniques to boost the effectiveness and durability of this intricate network \cite{yang2019hybrid}.

In recent times, there has been a significant increase in research focusing on electrical power systems \cite{wang2018research}, covering areas such as power system planning \cite{li2021power}, forecasting wind speed and power \cite{tsao2021center,silva2021novel}, predicting load \cite{wu2021support,singh2019hybrid}, and modeling energy demand and prices \cite{tang2018randomized,zhang2021multi}.

Several prior research investigations indicate that the usage of electricity can result in an increase in domestic production (GDP) growth, as noted by Waheed et al. \cite{waheed2019survey}. Several studies have delved into the complex correlation between energy consumption, carbon emissions, and economic advancement.
These investigations employ diverse methodologies to estimate causality, examining the interplay between these variables at individual and aggregate levels. However, conventional causality models delineate a directional relationship between variables; despite this, a consensus is yet to be reached in formulating definitive policies. The unresolved question in the literature pertains to whether distinctions exist between restrictive and more permissive policies, contributing to the ongoing debate surrounding policy implications.

The structure of the paper is as follows. Section \ref{sec2} provides a concise overview of causality in machine learning. Section \ref{sec3} outlines the proposed methodology and describes the data used. Lastly, Section \ref{sec4} presents the experimental findings.

\section{\textbf{Literature Review}}
\label{sec2}
The connection between electricity consumption, $\text{CO}_2$ emissions, and economic growth is vital across multiple sectors. This connection is particularly significant in economic policy formulation, requiring a nuanced strategy in managing both overall and sector-specific electricity usage alongside comprehensive environmental planning at national and global levels. The inherent variability in electricity consumption and carbon emissions across different sectors underscores the need for tailored strategies, acknowledging that each sector operates at distinct electricity consumption levels and emits carbon disparately. As a result, the scholarly interest in exploring and understanding these relationships has spawned many works considering diverse periods, indicators, and sets of countries and employing a spectrum of econometric techniques \cite{ang2008economic,rauf2018structural}.

Despite numerous studies and varied methodological approaches utilized, the results obtained display differences, leading to unique policy implications regarding the connections among $\text{CO}_2$ emissions, energy consumption, and economic growth \cite{waheed2019survey}. Hence, current research in this domain lacks definitive direction on the policy measures a country can undertake to enforce environmental regulations without compromising economic growth sustainability.

Similarly, further research has concentrated on clarifying the connections among $\text{CO}_2$ emissions, energy consumption, and the electricity market, utilizing a consistent methodological approach across their analyses \cite{Tiwari2021}. However, despite extensive research efforts, a consensus is yet to be reached concerning the theories relating to carbon emissions, energy consumption, and their connection with the electricity market. Moreover, the lack of clarity in developing concrete policies continues to persist. While classical causality models may offer directional insights between variables, the unresolved question in the literature revolves around the choice between implementing aggressive or more gradual policies. As a result, the utilization of intelligent causal treatment models in economics presents itself as a promising approach to addressing these challenges. Similarly, the application of neural networks in causal discovery has led to the adoption of specific deep learning models, such as the Balancing Counterfactual Regression \cite{Johansson2016} and the TARnet developed by Shalit et al. \cite{Shalit2016}. This concise literature review indicates a growing interest in applying neural networks in causality, prompting the exploration of novel architectures designed to effectively handle confounding variables and address the limitations inherent in traditional econometric models.

Causal Machine Learning is a significant advancement, facilitating the transition from predictive to prescriptive modeling. This progression represents a significant strategic advancement, enabling the anticipation of outcomes and the proactive recommendation of actions based on a thorough comprehension of causal connections inherent in the data. Predictive models typically operate by utilizing known input variables ($X$) to predict unknown output variables ($Y$), estimating the conditional expectation function $E(Y|X)$. However, this approach is inherently passive, providing limited control over outcomes. For instance, predicting the sales of a particular product in a store or forecasting inventory shortages are typical applications of predictive modeling. Conversely, Causal Machine Learning introduces an additional controllable treatment variable ($T$) alongside the input and output variables ($X$ and $Y$). This framework enables estimating causal relationships between the treatment and the output within the context of the input variables, denoted as $E(Y|X, T)$. The focus shifts towards actively influencing outcomes by identifying optimal treatments. For example, when predicting product sales, contextual factors such as historical sales data ($X$) and controllable treatment effects like pricing, inventory management, or marketing strategies ($T$) are considered \cite{Karmakar2023}.

CausalML is an open-source advanced Python package that provides a wide range of uplift modeling and causal inference techniques. It utilizes state-of-the-art machine learning algorithms to deliver comprehensive solutions.
CausalML is undergoing rigorous maintenance and development, spearheaded by the dedicated Uber development team. This team, comprised of Huigang Chen, Jeong-Yoon Lee, Totte Harinen, Zhenyu Zhao, and Mike Yung, plays a major role in advancing and refining the methodologies within the realm of CausalML. Conventional causal analysis techniques, like employing t-tests in randomized experiments (A/B testing), have demonstrated efficacy in gauging the Average Treatment Effect (ATE) of a particular treatment or intervention. Nevertheless, there arises a clear necessity and usefulness in various real-world situations to gauge these effects with greater detail and precision. CausalML offers a uniform interface that empowers users to calculate the Conditional Average Treatment Effect (CATE) or Individual Treatment Effect (ITE) from both observational and experimental datasets. This is also called the Heterogeneous Treatment Effect. This nuanced approach enables a more detailed understanding of the impact of interventions, contributing to a more precise and insightful analysis of diverse applications. In the context of CausalML, an essential modeling technique that is available to us is uplift modeling. Uplift modeling represents an advanced approach to causal learning aimed at estimating the individual treatment effect within experimental settings. This methodology enables users to uncover the subtle effects of a treatment, like a targeted marketing campaign, on an individual's behavior by leveraging experimental data \cite{Chen2020}.
\begin{figure*}[t]
  \begin{center}
  \includegraphics[width=8in]{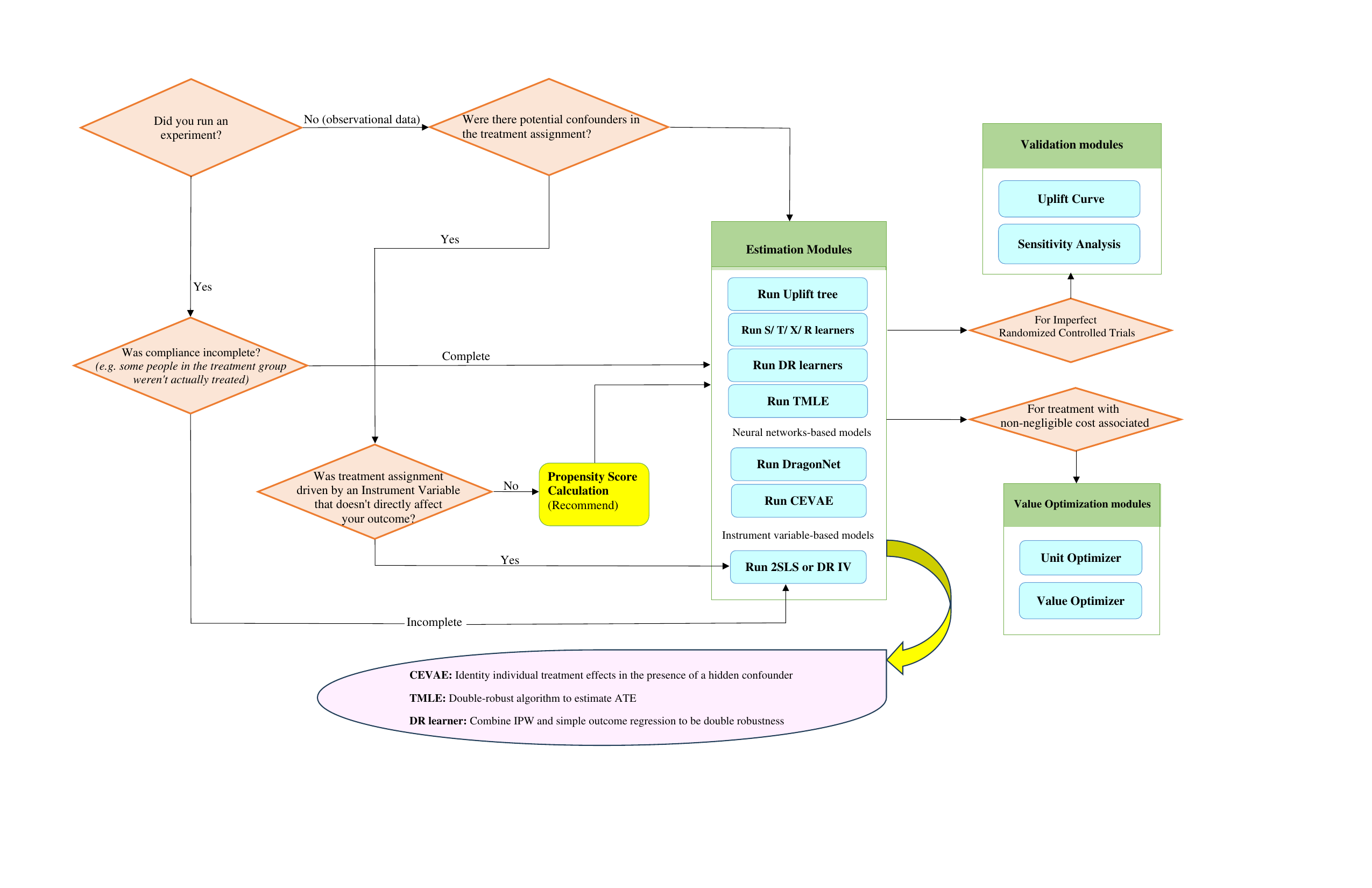}\\
  \vspace{-40pt} 
  \caption{Comprehensive Decision Guide Flowchart for Algorithm Selection.(Source; SIGKDD international conference, 2021, Introduction to CausalML, Page 41, 42)}\label{fig1}
  \end{center}
\end{figure*}

Uplift models, which are a subset of machine learning techniques, aim to compute the CATE from observational data. These models seek to forecast the additional value gained or "lift" generated by a specific treatment. They are broadly categorized into two groups: Direct Uplift Estimation methods and Meta-learners. Meta-learners encompass various types, including the single model method (S-learner), Transformed Outcome, and the Two-model approach (T-learners and X-learners), as well as Double Debiased Machine Learning (R-learner). The Two-model approach further differentiates between X-learners and T-learners. Conversely, Direct Uplift Estimation methods include Tree-based approaches (like Forests and Uplift Trees, Forests and Causal Trees), Instrumental Variables methods (such as Doubly Robust IV, 2SLS, Deep IV, and Ortho IV), and Neural Network methods (like IV, and Ortho IV), and Neural Network methods (like CEVAE and Dragonet) \cite{Karmakar2023}. This comprehensive taxonomy provides a structured overview of the diverse methodologies employed in uplift modeling. Fig. \ref{fig1} delineates a brief roadmap for the optimal algorithm selection tailored to a specific dataset. In the subsequent section of the paper, we meticulously elucidate the application of the meta-learner algorithm to our dataset, providing a comprehensive overview of the intricacies inherent in their operation. This exposition aims to convey a profound understanding of the applied algorithms, shedding light on their functionality and relevance within the context of our study.

\section{\textbf{Methodology}}
\label{sec3}
First, let us describe the causal effects based on the Rubin causal model (RCM), which was introduced by Rubin. We decided to use it as an example. 

Suppose researchers seek to appraise the causal influence of a new teaching method on students' mathematical performance. Now, there are two ways these students can be chosen. The first would be that they must be chosen by researchers randomly, and the second could be that they select the traditional or the new methods of teaching voluntarily. Now, we are confronted with two alternatives: 1) Random Assignment Mechanism and 2) Non-Random Assignment Mechanism. Please refer to King, Imbens, McKenzie, and Ridder (2008) for a better comprehensive understanding.
\subsection {Random Assignment Mechanism}
At first, to rigorously investigate this classification, we study each of them separately. Suppose they opt for a randomized controlled trial (RCT), employing random assignment to ensure a robust experimental design and mitigate potential biases.
The initial component of the Rubin Causal Model (RCM) centers around the concept of potential outcomes. For each student $i$, where $i =1, …, N$, the two potential outcomes are represented as $Y_i (0)$ and $Y_i (1)$.

$Y_i (0) \text{ is }$ The potential outcome for students $i$ if they do not receive the treatment, new method, (control group). $Y_i (1) \text{ is }$ The potential outcome for student $i$ under the condition of receiving the treatment (treatment group).
The causal effect ($\tau_i$) for a student $i$ is characterized as the difference between the potential outcomes under control and treatment conditions:
\begin{equation}
\tau_i = Y_i{(1)} - Y_i{(0)}.
\end{equation}

Population Average Treatment Effect (PATE):
PATE considers the entire population \cite{imbens2009recent}, not just those who receive the treatment. It considers the average effect of treatment for the entire population, whether they are treated or not, as follows:
\begin{equation}
\tau_{\text{PATE}} = \mathbb{E} [Y_i{(1)} - Y_i{(0)}],
\end{equation}
Where $\mathbb{E}$ is the expected value.

\vspace{12pt}

In this context, $\mathbb{E}$ signifies the expectation, and the average causal effect denotes the mean disparity between the outcomes of individuals who received treatment and those who did not in the population. To sum up, ATE concentrates on the average treatment effect among the subset of students who underwent treatment (new teaching method). In contrast, PATE assesses the average treatment effect across the entire population, considering both treated and untreated students.
After a specified period, the researchers measure students' performance in a standardized math test as the outcome variable. Any discernible variances in performance between the control and treatment groups can be ascribed to the implementation of the new teaching method, given that random assignment aids in mitigating the influence of confounding variables.

We will use the notation $Y_i$ to represent the observed outcome for student $i$, and $Y$ will be an $\mathbf{N}$-vector where the $i$-th element corresponds to $Y_i$.

\vspace{12pt}

The preceding discussion suggests that:
\begin{equation}
\begin{aligned}
Y_i &= Y_i(W_i) \\
    &= Y_i(0) \cdot (1 - W_i) + Y_i(1) \cdot W_i \\
    &= 
    \begin{cases} 
      Y_i(0) & \text{if } W_i = 0 \\
      Y_i(1) & \text{if } W_i = 1,
    \end{cases}
\end{aligned}
\end{equation}
where $W_i$ indicates whether student $i$ received the new method, with $W_i = 0$ if student $i$ did not, and $W_i = 1$ if individual $i$ received.

Researchers \cite{imbens2009recent} concentrate on the average causal effect conditional on the covariates in the sample,
\begin{equation}
\tau_{\text{CATE}} = \frac{1}{N} \sum_{i=1}^{N} \mathbb{E}[Y_i(1) - Y_i(0) | X_i].
\end{equation}
It evaluates the mean contrast between potential results among students who received treatment and those who did not, considering a group of covariates $X_i$.

This equation calculates the average impact of treatment, taking into account the subtle differences in treatment effects across various levels of covariates. Exploring the CATE enables researchers to discern the nuanced variations in treatment effects within subgroups delineated by the covariates.

In practical terms, you would estimate this by obtaining observed outcomes ($Y_i$) for students who received the treatment, new method ($W_i = 1$), and those who did not ($W_i = 0$). Then, conditional on the covariates $X_i$, you calculate the average outcome difference between treated and untreated students.
This formula is commonly used in observational studies and practical causal inference to account for potential heterogeneity in treatment effects across different characteristics. It helps researchers move beyond the average treatment effect to understand how the treatment effect varies based on individual characteristics.
The challenge in causal inference lies in the fact that just only one of the potential outcomes can be observed for each individual, determined by whether they receive the treatment (new method) or not. The unobserved potential outcome is commonly termed the \enquote{counterfactual} encapsulating what would have transpired under the alternative, unselected treatment condition.
When student $i$ is selected for the new method, we observe the realized outcome $Y_i(1)$, while the unobserved outcome $Y_i(0)$ serves as a counterfactual for what could have happened if they had not participated. Conversely, if student $i$ is not selected, we observe $Y_i(0)$, and $Y_i(1)$ becomes the counterfactual outcome representing what might have occurred if they were selected to be part of the test.

\subsection {Non-Random Assignment Mechanism}

Now, let us delve into a scenario where the feasibility of random assignment is compromised, necessitating researchers to pivot toward a non-random assignment mechanism. Faced with logistical constraints, the researchers opt to implement the new teaching method in schools that willingly adopt it. Meanwhile, those schools refraining from adopting the new method constitute the control group. Schools autonomously categorize themselves into either the treatment or control group based on their voluntary decision to adopt the new teaching method. This non-random assignment mechanism introduces the potential for selection bias, as schools possessing specific characteristics may exhibit a higher likelihood of adopting the new method. After the implementation period, the researchers measure students' performance in treatment and control groups.

\subsection {Simulations}
Understanding and accounting for the assignment mechanism is crucial in drawing valid causal inferences. The aim is to confirm that any detected discrepancies in outcomes between the control and treatment groups can be confidently ascribed to the treatment itself and not to biases introduced by the assignment process. Random assignment is considered the gold standard for achieving this goal. Still, when randomization is not possible, Non-random assignment introduces the challenge of potential confounding, as the treatment and control groups may differ systematically in ways not accounted for in the analysis. It is important to note that while these concepts provide a theoretical framework for expressing causal effects, estimating them in practice requires careful study design, and statistical methods such as matching, instrumental variables, propensity score analysis, and regression adjustment are applied to mitigate the effect of potential confounding factors and calculate causal effects in observational studies.
This section examines our CausalML algorithm as it is applied to our specific dataset. We meticulously review pertinent frameworks and definitions drawn from the academic literature to facilitate a contextual understanding of our analysis. 
As mentioned, we employed a meta-learner algorithm using the CausalML package in Python. Here, we want to explain this methodology, which is supported as one of the algorithms in CausalML developed so far. A meta-algorithm, known as meta-learning, serves as a conceptual framework for estimating the CATE through various machine-learning estimators. A meta-algorithm is distinguished by its utilization of either a single base learner that includes the treatment indicator as a feature (such as the T-learner) or the employment of multiple separate base learners assigned to each of the treatment and control groups (such as the S-learner, X-learner, and R-learner).

\vspace{12pt}

\subsubsection{Efficiency Bounds}
The calculation of confidence intervals for estimates of the average treatment effect involves utilizing a lower-bound formula, denoted as formula 6 \cite{imbens2009recent}. Before delving into particular estimation methods, examining the insights we can obtain regarding the parameters of interest is beneficial solely based on the assumption of strong ignorability of treatment assignment. This analysis is conducted without incorporating specific assumptions about functional forms or distributions. To achieve this, some extra notations are required.

Let,
\begin{equation}
\begin{aligned}
\sigma_0^2(x) \triangleq \text{Var}(Y_i(0) | X_i = x) \\
\sigma_1^2(x) \triangleq \text{Var}(Y_i(1) | X_i = x),
\end{aligned}
\end{equation}
where represent how the potential outcomes are spread based on the given covariates.
Hahn \cite{hahn1998role} finds the smallest possible expected errors for estimators that get more accurate with more data ($\sqrt{N}$-consistent estimators) for $\tau_{\text{PATE}}$, represented by
\begin{equation}
V_{\text{PATE}} = \mathbb{E} \left[ \frac{\sigma_1^2 (X_i)}{e(X_i)} + \frac{\sigma_0^2 (X_i)}{1 - e(X_i)} + (\tau(X_i) - \tau)^2 \right].
\end{equation}

\vspace{12pt}

\subsubsection{Meta learners}
  Utilizing the Neyman-Rubin (NR) model or potential outcomes framework, the variables can be defined as follows: $Y_i(0)$ represents the potential outcome of unit $i$ if it were assigned to the control group. In contrast, $Y_i(1)$ signifies the potential outcome if unit $i$ were assigned to the treatment group. $Y$ denotes the continuous outcome, $W_i$ signifies the treatment-assignment indicator, $W_i \in \{0,1\}$, and $X_i$ is a covariate or feature vector. In meta-learning methodologies, the treatment space must be discrete. This means that when meta-learning algorithms are used to estimate treatment effects, the treatments or interventions are represented by distinct categories or options. In other words, the treatments cannot be continuous or infinitely variable; instead, they are defined as discrete choices or categories. This requirement is specific to how meta-learning algorithms are structured and utilized to estimate treatment effects.

\vspace{12pt}

\subsubsection*{\textbf{S-learner}}
The S-learner computes the treatment effect by utilizing a lone machine-learning algorithm in the subsequent manner:

\vspace{12pt}

Step 1:
Within the S-learner algorithm, the treatment indicator variable $W$ is included as a feature, receiving no preferential treatment or unique distinction compared to other features. Consequently, estimate the collective response $\mu(x)$, considering covariates $X$, using a base learner (such as a regression algorithm or supervised machine learning) applied to the entire dataset.

\begin{equation}
\mu(x, \omega) = \mathbb{E}[Y | X = x, W = \omega]
\end{equation}

\vspace{12pt}

Step 2:
Specify the CATE estimate as: 
\begin{equation}
\hat{\tau}(x) = \hat{\mu}(x, W = 1) - \hat{\mu}(x, W = 0).
\end{equation}

\vspace{12pt}

Incorporating the propensity score into a statistical model can help mitigate bias due to confounding induced by regularization \cite{hahn2020bayesian}. Regarding the example of a new teaching method for students, suppose that students are either assigned to take the new teaching method (treatment group) or not (control group). However, some underlying factors (covariates) might influence the assignment to the treatment group and the outcome. This situation is known as confounding. When we use a model to analyze the outcomes, regularization techniques can be applied to prevent overfitting or enhance generalization. However, regularization can inadvertently introduce bias if it does not correctly account for the confounding factors. To tackle this issue, we integrate the propensity score into our model. The propensity score represents the likelihood of receiving the treatment based on the covariates. By integrating this score, we can ensure balance in covariates among the control and treatment groups. This helps reduce the bias caused by regularization as the model becomes more aware of the likelihood of receiving the treatment, leading to a more precise assessment of the treatment effect.

If there are substantial differences in covariates between the control and treatment groups, a single linear model may not adequately capture the distinct related characteristics and smoothness of features for both groups \cite{alaa2018limits}.

Once more, let us revisit our example involving a new teaching method. Students are haphazardly assigned to either the control group or the treatment group (receiving the new teaching method). Now, consider that the participants in the treatment and control groups differ significantly in intelligence, gender, educational background, coming from what type of family class (low-income, middle-class, high-income families), and other characteristics (known as covariates). In this scenario, applying a single linear model to predict the outcomes (scores in the math test) based on these covariates might oversimplify the situation. The linear model assumes that the relationship between covariates and outcomes is similar for the treatment and control groups. However, if the groups are so different, a single model may struggle to capture the unique patterns and variations in outcomes for each group. For instance, the treatment group might respond differently to the new teaching method based on specific covariates, and the control group might have distinct patterns unrelated to the treatment. Applying a single linear model may not effectively capture these differences.

More sophisticated modeling approaches can address this limitation, such as separate models for the treatment and control groups (like the T-learner, X-learner, R-learner, or other heterogeneous treatment effect models). These approaches allow for flexibility in capturing the diverse effects of the treatment on different subgroups within the study population.

\vspace{12pt}

\subsubsection*{\textbf{T-learner}}
The T-learner \cite{kunzel2019metalearners} involves two steps, outlined as follows:

Step 1:
The control response $\mu_0(x)$ can be computed using a base learner, a supervised machine learning, or a regression estimator. This is accomplished by leveraging observations in the control group, denoted as $f(X_i, Y_i) | W_i = 0$. Subsequently, the treatment response $\mu_1(x)$ can be estimated using a potentially distinct base learner, utilizing treated observations.

\begin{equation}
\begin{aligned}
\mu_0(x) &= \mathbb{E}[Y(0) | X = x] \\
\mu_1(x) &= \mathbb{E}[Y(1) | X = x]
\end{aligned}
\end{equation}

\vspace{12pt}

Step 2:
Specifying the CATE estimate as:
\begin{equation}
\hat{\tau}(x) = \hat{\mu}_1(x) - \hat{\mu}_0(x).
\end{equation}

\vspace{12pt}

\subsubsection*{\textbf{X-learner}}
The X-learner \cite{kunzel2019metalearners} serves as an expansion of the T-learner and comprises three steps, delineated as follows:

\vspace{12pt}

Step 1:
Compute the mean outcomes, represented as $\mu_0(x)$ and $\mu_1(x)$, for both cohorts utilizing any supervised machine learning or regression algorithm. These algorithms, referred to as the base learners in the initial step, play a role in the estimation procedure.

\begin{equation}
\begin{aligned}
\mu_0(x) &= \mathbb{E}[Y(0) | X = x] \\
\mu_1(x) &= \mathbb{E}[Y(1) | X = x]
\end{aligned}
\end{equation}

Step 2:
Estimate the treatment effects for individuals (e.g., students) in the treated group using the control-outcome estimator. Similarly, the treatment effects for individuals in the control group were predicted using the treatment outcome estimator, referred to as the imputed treatment effects.

\begin{equation}
\begin{aligned}
D_i^1 &= Y_i^1 - \hat{\mu}_0(X_i^1) \\
D_i^0 &= \hat{\mu}_1(X_i^0) - Y_i^0
\end{aligned}
\end{equation}

\vspace{12pt}

Utilize any supervised machine learning or regression technique(s) to estimate $\tau(x)$ in two ways: employ the imputed treatment effects as the outcome variable in the treatment group to derive $\tau_1(x)$, and similarly in the control group to derive $\tau_0(x)$. These supervised machine learning or regression algorithms are referred to as the base learners in the second step.

\begin{equation}
\begin{aligned}
\tau_0(x) &= \mathbb{E}[D_0 | X = x] \\
\tau_1(x) &= \mathbb{E}[D_1 | X = x]
\end{aligned}
\end{equation}

\vspace{12pt}

Step 3:
Specify the CATE estimate as a weighted average of the two estimates obtained in step 2:
\begin{equation}
\tau(x) = g(x) \tau_0(x) + (1 - g(x)) \tau_1(x),
\end{equation}
Where $\tau(x)$ is the overall estimator,
$g(x)$ is a weight function that assigns a weight between 0 and 1 to each observation,
$\tau_0(x)$ and $\tau_1(x)$ are two different estimators for the treatment effect.

It suggests estimating the propensity score for $g$, mentioned as $\hat{e}$. As told before, it is highly used in observational studies to compute the probability of receiving treatment based on observed covariates. This approach assumes that the weight function $g$ is a function of the estimated propensity score: $g = \hat{e}$.

It means that it might make sense to set $g = 1$ or $g = 0$ in certain situations:
\begin{itemize}
    \item If the number of treated units is huge compared to the number of control units, setting $g = 1$ may be reasonable.
    \item If the number of treated units is minimal compared to the number of control units, setting $g = 0$ may be reasonable.
\end{itemize}

Some estimators may allow for estimating the covariance matrix of $\tau_0(x)$ and $\tau_1(x)$. The choice of $g$ can be made to minimize the variance of $\tau(x)$. Minimizing the variance is a common goal in statistical estimation, as it leads to more precise and reliable estimates.

In summary, the description discusses the flexibility in choosing the weight function $g$ in the weighted combination of estimators. This choice can be based on the estimated propensity score, set to specific values in certain scenarios, or even optimized to minimize the variance of the overall treatment effect estimator. The target is to enhance the accuracy and reliability of the treatment effect estimation in observational studies.

\vspace{12pt}

\subsubsection*{\textbf{R-learner}}
The core concept of the R-learner \cite{nie2017quasi} is to create a formula that represents how treatment effects differ across individuals. Then, a unique formula called a loss function (R-loss function) can be devised to catch these differences. The target is to find the best solution for this formula, considering both accuracy and simplicity, to get reliable estimates of treatment effects.

In R-learner, a crucial part is the R-loss function used for estimating treatment effects through cross-validation. 

We suppose that we have $n$ examples, each with features $(x_i)$, an observed outcome $(y_i)$, and a treatment assignment $(t_i)$. The potential outcome $(Y_i)$ is what we would have observed if the treatment assignment were $0$ or $1$ \cite{nie2017quasi}, denoted as $Y_i(t_i)$. Assuming that treatment assignment is random once we consider the features $(x_i)$ (this is called unconfoundedness), we aim to estimate the CATE function \cite{rosenbaum1983central}, denoted as
\begin{equation}
\tau(x) = \mathbb{E}\{Y(1) - Y(0) | X = x\}.
\end{equation}

\vspace{12pt}

To do this, the treatment propensity and conditional surfaces can be used, representing how likely a unit is to receive treatment and the expected outcomes based on the features. In simpler terms, we're trying to figure out the average effect of treatment compared to no treatment, considering features and ensuring randomness in treatment assignment. We use the treatment propensity and conditional surfaces to help us make these estimations \cite{nie2017quasi}:

Estimating treatment propensity scores: 
\begin{equation}
\text{\textit{Pr}} = \text{Probability}\quad e^*(t) = \text{\textit{Pr}}(T = 1 | X = x)
\end{equation}

Note that $*$ superscripts signify unknown population parameters.

Estimating conditional response surfaces:
\begin{equation}
u_t^* = \mathbb{E}\{Y(t) | X = x\} \quad \text{for } t \in \{0, 1\}
\end{equation}

\vspace{12pt}

Conditional Mean Outcome:
\begin{equation}
m^*(x) = \mathbb{E}(Y | X = x) = \mu^*(0)(X_i) + e^*(X_i) \tau^*(X_i)
\end{equation}

\vspace{12pt}

Re-writing the CATE using conditional mean outcome function gives the below equation in the output \cite{nie2017quasi};

\begin{equation}
Y_i - m^*(x_i) = (W_i - e^*(X_i)) \tau^*(X_i) + \epsilon_i,
\end{equation}
where $\epsilon_i$ is the error term, capturing unobserved factors or noise.
The formula 19 is called Robinson’s transformation \cite{robinson1988root}. The term \enquote{transformation} suggests that it involves converting or altering some variables or quantities in a specific way. Robinson's transformation incorporates concepts and methodologies from modern machine learning. This means that it uses sophisticated machine learning algorithms to enhance the accuracy and flexibility of treatment effect estimation. The term \enquote{flexibility} suggests that this method is adaptable and can handle a variety of scenarios or data characteristics. This transformative approach is designed to compute parametric elements within partially linear models, attracting substantial attention in recent times.

Athey, Tibshirani, and Wager \cite{athey2019generalized} (2019) use this approach to cultivate a causal forest resilient to confounding extends its application in devising G-estimation for periodical trials \cite{richardson2014causal}, while Chernozhukov \cite{chernozhukov2018double} showcase it as a prominent illustration of the effective utilization of machine learning methods for estimating nuisance components in semiparametric inference. 

So, the modeled CATE based on the model and data formula \cite{richardson2014causal}:

\begin{equation}
\begin{aligned}
\tau^*(\cdot) &= \underset{\tau}{\text{argmin}}  \bigg\{ \mathbb{E} \Big[ \big\{ Y_i - m^*(X_i) \\
&\quad - \left( W_i - e^*(X_i) \right) \tau(X_i) \big\}^2 \Big] \bigg\}.
\end{aligned}
\end{equation}

\vspace{12pt}

The true CATE can be obtained for the entire population from the formula below. 

\begin{equation}
\begin{aligned}
\tau^*(\cdot) &= \underset{\tau}{\text{argmin}}  \Bigg( \frac{1}{n} \sum_{i=1}^{n} \bigg[ Y_i - m^*(X_i) \\
&\qquad - \left( W_i - e^*(X_i) \right) \tau(X_i) \bigg]^2 + \Lambda_n \{\tau(\cdot)\} \Bigg).
\end{aligned}
\end{equation}

Think of $\Lambda_n(\tau(\cdot))$ as a way to prevent the $\tau(\cdot)$ function from becoming too complex. This control, called regularization, can be direct, like in penalized regression, or more subtle, such as in a well-designed deep neural network. The challenge is that in practice, we typically don't have access to the weighted main effect function $m^*(x)$ and often lack knowledge of the treatment propensities $e^*(x)$. As a result, formula 21, mentioned earlier, is not something we can practically use.

Let’s concrete it with an example;

Think of $\tau^*(\cdot)$ as the \enquote{Weather App Prediction} for tomorrow's temperature. The app uses various data and a model to estimate the temperature, giving you its best guess. However, it might not always perfectly predict the actual temperature. On the other hand, the \enquote{True Temperature} is what you would measure with the most accurate thermometer available. This would be tomorrow's real, exact temperature – something you'd love to know but might not have the perfect tool to measure. With these basics in mind, we're now examining a specific type of two-step estimator \cite{chernozhukov2018double} that involves cross-fitting. It includes two steps as follows:

Step 1: Learning process
Split the data into equally sized parts $Q$ (usually 5 or 10 folds). Use $q(\cdot)$ to map each sample index ($i = 1, \ldots, n$) to one of these $Q$ folds. Then, perform cross-fitting over these $Q$ folds to fit $\hat{m}$ and $\hat{e}$ using methods optimized for the best predictive accuracy.

Step 2: Optimizing process
Compute treatment effects utilizing a modified form of equation 21, where predictions such as $\hat{e}^{-q(i)}(X_i)$ indicate making predictions without incorporating the data fold to which the $i$-th training example belongs.

So the loss function (R-loss) can be achieved as follows:
\begin{equation}\label{eq:my_equation}
\begin{aligned}
\hat{L}_n\{\tau(\cdot)\} &= \frac{1}{n} \sum_{i=1}^{n} \left[ Y_i - \hat{m}^{(-q(i))}(X_i) \right. \\
&\quad \left. - \{W_i - \hat{e}^{(-q(i))}(X_i)\} \tau(X_i) \right]^2.
\end{aligned}
\end{equation}

\vspace{12pt}

Now, the mathematical representation of CATE is: 
\begin{equation}
\hat{\tau}(\cdot) = \underset{\tau}{\text{argmin}} \left[ \hat{L}_n(\tau(\cdot)) + \Lambda_n(\tau(\cdot)) \right].
\end{equation}

\vspace{12pt}

This methodology is called R-learner, paying homage to Robinson (1988) and underscoring the significance of residualization in this approach. Additionally, we denote the squared loss $\widehat{L}_n\{\tau(\cdot)\}$ as the R-loss for clarity and conciseness in our discussion \cite{nie2017quasi}.

\section{\textbf{Estimations}}
\label{sec4}
We used Colab to write and execute the project’s Python code. We used 16 GB RAM on the Colab notebook to run the code. You must install Python packages Causalml (0.14.1) and Econml (0.14.1), updated to the latest version when writing this paper, on the notebook environment in which you are writing your code. You can use the Python package ‘TensorFlow’ to install causalml and econml in your notebook. We executed the code for 100 times, and we suggest you run your final code for several times. In this project, we use different types of algorithms as the base learner for our meta-learners. The best algorithm depends on your dataset, like how the data is spread out and how big it is. You can pick from various algorithms like penalized methods, Bayesian Additive Regression Trees (BARTs), Random Forests (RFs), kernel

\begin{figure*}[t]
\centering
\includegraphics[width=7.1in]{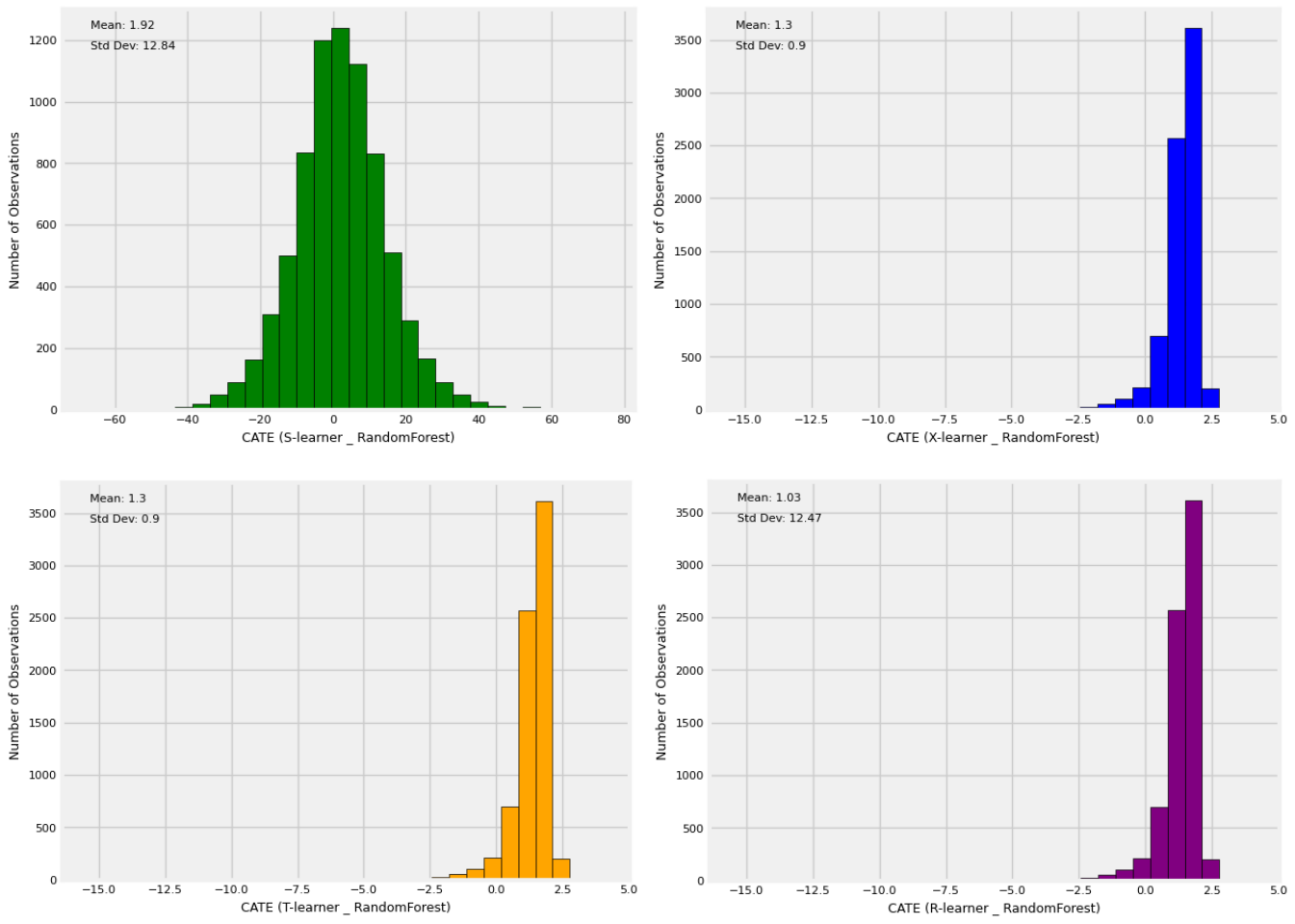}
\vspace{-40pt} 
\captionsetup{justification=centering}
\caption{CATE estimates using learners S, T, R, and X.}\label{fig2}
\end{figure*}
ridge regression, gradient boosting, and more. We recommend trying different methods in your meta-learners until you find the one that gives you the best result. We suggest that if you use a big dataset, use less expensive algorithms first.

\vspace{12pt}

Please note:

•	To prevent biased estimations, the treatment and control groups' sizes must be unequal.

•	The optimal treatment condition depends on your research goals and policy interests.

•	Defining a binary treatment condition in an observational dataset necessitates addressing selection bias. Matching techniques, propensity score weighting, or instrumental variables might be necessary to validate the CATE estimate's causal interpretation.

Visualized in Fig. \ref{fig2}, the CATE values for learners T and X converge at 1.3, signifying a similar treatment impact. Learner S demonstrates the highest CATE at 1.92, indicating a more substantial treatment effect. Conversely, learner R presents the lowest CATE at 1.03, suggesting a comparatively minor impact. We know that by increasing consumers' demand for electricity, electricity companies should generate more electricity, increasing $\text{CO}_2$ intensity. So, policymakers, especially environmental activists, try to define a rebate on electricity bills as a possible solution to encourage consumers to consume less electricity to protect our environment. Our datasets are collected from a region called West Virginia in Ireland. This dataset includes 37868 records. The causal scenario entangles four discrete covariates: wind speed, temperature, electricity price, and system load. In our analysis, the control group is assigned to 70 percent of records, and the treatment group is assigned to the other 30 percent of records.

 In our causal scenario to estimate CATE for the outcome, decrease in $\text{CO}_2$ intensity, we define two treatment conditions as follows;

 The binary treatment condition: Provide a 50 percent rebate on electricity bills for the treated group when the average $\text{CO}_2$ intensity of electricity generation dips below a target level (e.g., below a specific grams per kilowatt-hour value is considered as the load on the electrical system). The consumers in the treatment group are selected randomly from all the records.  The control group receives no such incentive.
CATE Estimate: This would capture the average price change for the treated group when generating electricity with lower carbon emissions. 
CATE, or Conditional Average Treatment Effect, exhibits distinct variations across different analyses, influenced by many factors. 
The final CATE is susceptible to various elements, including how the problem is defined, the contextual nuances surrounding it, and the specific goals aimed at achieving it. Depending on these considerations, the interpretability of CATE becomes particularly nuanced and insightful.

\begin{figure}
\centering
\includegraphics[width=3.0in]{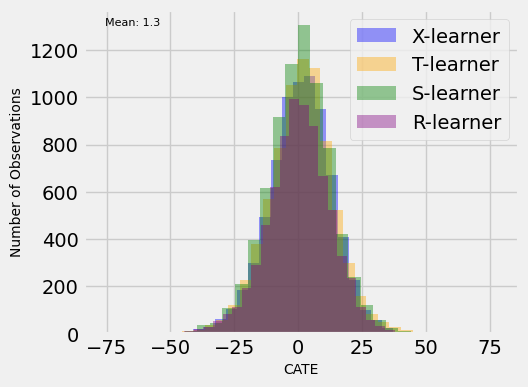}
\caption{Overlapped histogram of distributions of CATE according to applied algorithms.}
\label{fig3}
\end{figure}

 The histogram in Fig. \ref{fig3} shows an overlapped histogram of learners for estimating the CATE (Conditional Average Treatment Effect), which typically refers to a visualization that displays the distribution of estimated treatment effects from different machine learning models or algorithms. Based on the established goals in our causal scenario, it can be inferred that implementing a 50 percent discount on electricity bills may result in a 1.3 unit increase in $\text{CO}_2$ intensity (measured in grams per kilowatt-hour) based on the studied population.

\begin{figure}
\centering
\includegraphics[width=3.4in]{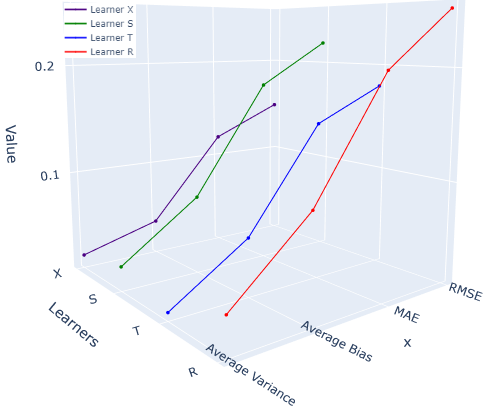}
\caption{RMSE, MAE, Variance, and Bias estimates for the applied meta-learners.}
\label{fig4}
\end{figure}

In Fig. \ref{fig4}, the computed Root Mean Square Error (RMSE) values provide valuable insights into the learners' performance. Noteworthy is the exemplary efficacy exhibited by learner X, boasting an impressively low RMSE of 0.15, and learner T, with a commendable RMSE of 0.18. In contrast, learners R and S lag, showing comparatively higher RMSE values of 0.25 and 0.22, respectively. This disparity underscores the varying degrees of proficiency among the learners, emphasizing the significance of these findings in assessing performance.

The Mean Absolute Error (MAE) values provide insights into the performance of the learners R, T, S, and X in estimating heterogeneous treatment effects using meta-learners. MAE measures the average magnitude of errors in a set of predictions, with lower values indicating better performance. Achieved MAE = 0.2 for learner R indicates that, on average, the predictions made by learner R have an absolute error of 0.2. In other words, the difference between the predicted and actual treatment effects is, on average, 0.2.  The MAE= 0.12 achieved for learner X demonstrates the best performance among all the learners, with the lowest MAE of 0.12. This indicates that its predictions have the most minor average absolute error (0.12), making it the most accurate in estimating treatment effects compared to the other learners. 

The average variance values obtained for the learners R, T, S, and X are 0.04, 0.015, 0.03, and 0.02, respectively. These values provide insights into the variability or dispersion of the treatment effects estimated by each learner. The R learner has the highest average variance among all learners (0.04). A higher variance suggests that the estimated treatment effects from the R learner are more spread out or less consistent. This could indicate that the R learner is more sensitive to the data or less stable in its estimations than other learners. The T learner has the lowest average variance (0.015) among all learners. A lower variance indicates that the estimated treatment effects from the T learner are more consistent or less variable. This suggests that the T learner provides more stable and reliable treatment effect estimations than others.

In estimating heterogeneous treatment effects using different learners (R, T, S, and X), the average bias values provide insights into the accuracy and reliability of each learner's estimates. With an average bias value of 0.1, the R learner has the highest bias among all the learners. This suggests that the estimates produced by the R learner tend to deviate more from the true treatment effects than the other learners. Higher bias indicates a higher likelihood of systematic error in the estimates. The X learner has the lowest average bias value of 0.04 among the four learners. This indicates that the estimates produced by the X learner are the least biased among all the learners, implying higher accuracy and reliability in estimating treatment effects.

\vspace{12pt}

\section{\textbf{Conclusion}}
As the electricity market is considered one of the fields that can highly affect a country’s GDP, it can be expected to become more and more competitive. But along with its benefits and cost to deliver to the public by bringing harmful environmental effects, it is considered one of the social concerns, especially for the environmental activists and the countries that joined the Paris Agreement 2016. All these concerns can give value to pricing policies that policymakers can adopt in the electricity market. This study’s investigations show that the financial incentives in electricity pricing for the consumers to use less electricity do not necessarily lead to a cleaner environment by decreasing $\text{CO}_2$ emission by the electricity companies' electricity production. As this study’s analysis shows, it can be inferred that applying a 50 percent discount on electricity bills can increase the $\text{CO}_2$ intensity. Regarding the more covariates for future studies by the researchers, like consumer behavior or applying different methods can either give value to this study or bring contradictory comments. But the clear thing is that more studies are needed to be done by researchers with ecological concerns.

\ifCLASSOPTIONcaptionsoff
  \newpage
\fi

\bibliographystyle{IEEEtran}
\bibliography{main}

\vfill

\end{document}